\documentclass[runningheads]{llncs}
\usepackage{graphicx}
\usepackage[utf8]{inputenc}
\usepackage{color}
\usepackage{multirow}
\usepackage{amsmath}

\begin{document}
%

\title{Reciprocal Feature Learning via Explicit and Implicit Tasks in Scene Text Recognition\thanks{\small{Both H. Jiang and Y. Xu contributed equally. Email: chengzhanzhan@hikvision.com. Supported by National Key R\&D Program of China(Grant No. 2018YFC0831601).} }}

%
\titlerunning{Reciprocal Feature Learning via Explicit and Implicit Tasks in STR}
%
\author{
Hui Jiang\inst{1} \and
Yunlu Xu \inst{1}\and
Zhanzhan Cheng\inst{2,1} \and
Shiliang Pu\inst{1} \and
Yi Niu\inst{1} \and
Wenqi Ren\inst{1} \and
Fei Wu\inst{1} \and
Wenming Tan\inst{1}
}
\authorrunning{H. Jiang, Y. Xu, Z. Cheng et al.}
%
\institute{Hikvision Research Institute, China \and Zhejiang University, Hangzhou, China
}
\maketitle              

\begin{abstract}
	Text recognition is a popular topic for its broad applications. In this work, we excavate the implicit task, character counting within the traditional text recognition, without additional labor annotation cost. The implicit task plays as an auxiliary branch for complementing the sequential recognition. We design a two-branch reciprocal feature learning framework in order to adequately utilize the features from both the tasks. Through exploiting the complementary effect between explicit and implicit tasks, the feature is reliably enhanced. Extensive experiments on 7 benchmarks show the advantages of the proposed methods in both text recognition and the new-built character counting tasks. In addition, it is convenient yet effective to equip with variable networks and tasks. We offer abundant ablation studies, generalizing experiments with deeper understanding on the tasks. Code is available at https://davar-lab.github.io/publication.html.

\end{abstract}

\section{Introduction}
Scene text recognition (STR) has attracted great attention in recent years, whose goal is to translate a cropped scene text image into a string.
Thanks to the rapid development of deep learning, modern STR methods have achieved remarkable advancement on public benchmarks \cite{wrong}.
As the high precision acquired on most of the commonly used datasets, \textit{e.g.}, above 90\% or even 95\% accuracy on IIIT5K\cite{IIIT5K}, SVT\cite{SVT}, ICDAR 2003/2013\cite{IC13},
recent works tend to more specific problems instead of general representative ability, including irregular text \cite{aster}, low-quality imaging \cite{PlugNet}, attention drift \cite{FAN,DAN,DSAN}, efficiency \cite{ACE} and \textit{etc.}.
An emerging trend is to enhance the networks with additional branches \cite{FAN,RobustScanner,DSAN,OOV}, generating a tighter constraint on the original sequential task.
It can be roughly grouped in two categories: (1) Approaches introducing \textit{additional annotations or tasks} assisting the sole sequential recognition and (2) Approaches \textit{exploiting the original tasks} limited to the word-level annotations.

In the first group, using \textit{additional information from another task or detailed supervision} is intuitive and usually ensures the boosting performance.
For instance, in FAN \cite{FAN} and DSAN \cite{DSAN}, character-level position was imported for finer constraints on word recognition.
SEED \cite{SEED1} brought pre-trained language model supervised from wording embedding in Natural Language
 Processing (NLP) \cite{FastText} field to compensate the visual features.
However, all the above approaches bring in extra annotation cost, enlarging from word-level labels to character-level or needing source from other domains, \textit{e.g.}, language model.

Another group \textit{exploiting original tasks and supervision} without introducing extra supervision or tasks is meanwhile sprouting, such as multi-task learning\cite{joint} of variable networks (\textit{e.g.,} CTC-based \cite{CTC} and Attn-based \cite{ATTENTION1}\footnote[1]{CTC is short for Connectionist Temporal Classification, and Attn is short for Attention Mechanism in the whole manuscript, both of which are mainstreaming approaches in sequence-to-sequence problems.} models) sharing partial features.
Though expecting the joint learning of two branches for supplementary features on both sides, the improvement is not always satisfying \cite{FAN,GTC}.
{
As claimed in \cite{GTC}, CTC branch is promoted under the guidance of Attn, which turns into the main contribution in GTC approach that gets a more efficient and effective CTC model.
Conversely, the Attn branch cannot be likewise promoted with CTC, since the feature learning in Attn is harmed by the joint partner CTC loss.
In essence, the multi-task training seems like a trade-off between the features from the two branches. We attribute it to the similartaxis of dual-branch rooting in the same supervision, the original strong branch can not be enhanced any further.
\textit{In another word, the limitation of the multi-task learning in STR (without extra annotations) lies in the onefold word-level supervision.}
}
\begin{figure}[t]
	\includegraphics[width=\textwidth]{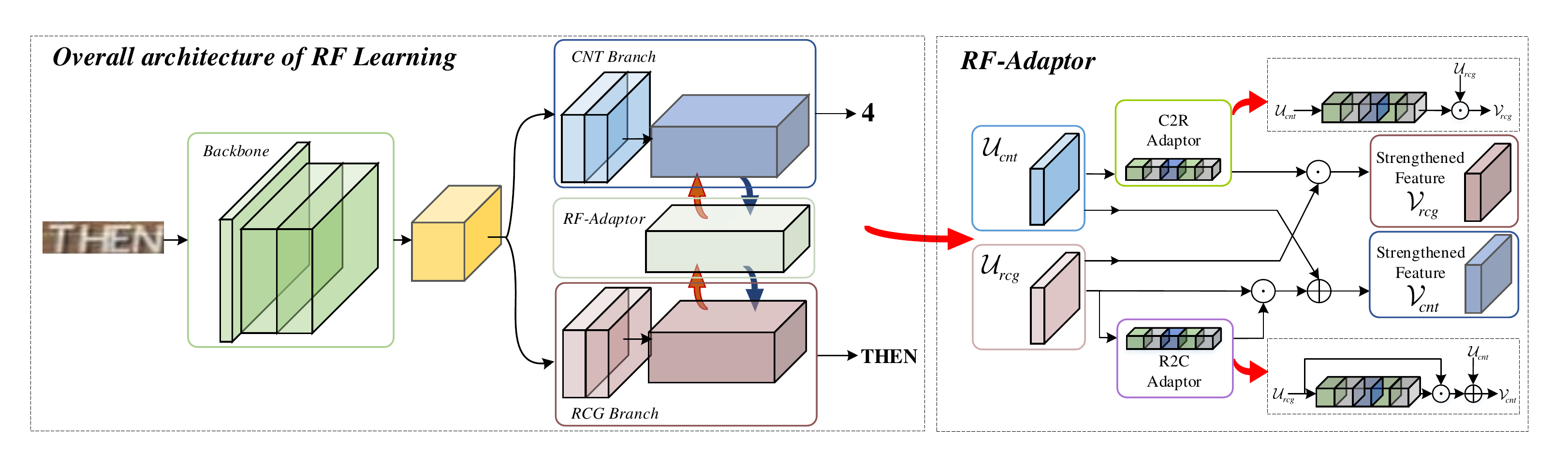}
	\caption{Overall architecture of the proposed network. Given an image input, the shared backbone extracts visual features, which then are fed into the RCG and CNT branch. The two branches are trained by respective supervision, and intertwined and mutually assisted through a novel RF-Adaptor.}
	\label{figure:architectore_comp}
\end{figure}
So far, what we are wondering is {\textit{`Can a strong recognizor be promoted with another task without extra annotation cost?'}}. Observing there is hierarchically implicit information in text itself, including tokens, semantics, text-length and the like, apart from text strings.
Obviously, the implicit labels above cannot cover complete information in the string, but we make a conjecture that it leads the network to learn its typical features differing from the usual text labels.
Regarding not bringing extra annotations or resources(\textit{e.g.,} tokens or semantics in text relies on extra corpus), we take the text-length as the additional supervision, generating a new task named \textit{character counting}. The total character numbers can be directly obtained from known words, not requiring any labor cost.

In this paper, we propose a novel text recognition framework, called Reciprocal Feature Learning (RF-L). It consists of two tasks.
The main task is a recognizor (denoted as RCG) outputting a target string.
And we also exploit the implicit labels to generate an auxiliary task named \textit{counting} (denoted as CNT) to predict the occurrence number of characters, as shown in Fig. 1.
The supervision of CNT can be inferred from RCG, but the training goals have some difference. The feature learned in CNT focuses more on the category-unaware occurrence of characters, while the RCG puts effort on discrepancy between types of classes and even the End-Of-Sentence(EOS) signals.
Intuitively, we hope to utilize the discrepancy and build relation between the two branches with separate tasks.
In this work, we propose a well-designed module called Reciprocal Feature Adaptor (RF-Adaptor). It transfers the complementary data from one branch to the other without any extra annotations, playing a role of assembling features and adapting to the task. Note that the RF-Adaptor allows bi-directional data flowing between both the tasks.
The main contributions can be summarized as:

(1) We dig the implicit task in the traditional STR, \textit{i.e.,} character counting, without any extra annotation cost. The counting network supervised by new exploited labels can be regarded as an auxiliary part in addition to the recognition task, facilitating positive outcomes. Also, we offer a strong baseline for the sole newly-built character counting task based on the existing STR datasets.

(2) We propose a multi-task learning framework called RF-L for STR through exploiting the complementary effect between two different tasks, word recognition and counting respectively. And the two tasks are learned in their own branch, in interaction to the other, via a simple yet effective RF-Adaptor module.

(3) The proposed method achieves impressive improvements in multiple benchmarks, not only in STR tasks but also in counting. The auxiliary network and the adaptor can be easily integrated into deep neural network with any other scene text recognition method, which boosts the single task via the proposed RF-L framework as verified in extensive experiments.

\section{Related Works}
\subsection{Scene Text Recognition}
Scene text recognition is an important challenge in computer vision and various methods have been proposed. Conventional methods were based on hand-crafted features including sliding window methods \cite{SVT}, strokelet generation \cite{strokelets}, histogram of oriented gradients descriptors \cite{su2014accurate} and \textit{etc.}. Recent years, deep neural network has dominated the area. Approaches can be grouped in \textit{segmentation-free} and \textit{segmentation-based} methods.

In the first \textit{segmentation-free} category,
Shi et al.\cite{CRNN} have brought recurrent neural networks for the varying sequence length, and the CTC-based recognition poses a milestone in OCR community, not only STR, but also the handwriting\cite{CTC} or related tasks \cite{handwritting}. Later works \cite{aster,FAN} applied attention mechanism in STR towards higher level of accurate text recognition. Recently, some specific modules appeared like transformation \cite{aster}, high-resolution pluggable module \cite{PlugNet}, advanced attention \cite{DAN} and enhanced semantic modeling \cite{SRN}. Although focused on different modules or handling with variable problems, their main body can be roughly classified into CTC-based \cite{CRNN,2dctc} and Attn-based methods \cite{aster,FAN,SRN,DAN}.
\textit{Segmentation-based} approaches are becoming another new hotspot \cite{Perspective,TextScanner}. They are usually more flexible than sequential decoders in the recognition of irregular text with orientation or curvation, and offer finer constraints on features, due to the detailed supervision.
However, these methods remain limited because they can not work with only sequence-level labels.
\subsection{Auxiliary Supervision and Tasks in STR}
As modern networks\cite{wrong,aster} have acquired high precision on most of the commonly-used datasets, \textit{e.g.}, above 90\% or even 95\% accuracy on IIIT5K\cite{IIIT5K}, SVT\cite{SVT}, ICDAR 2003/2013\cite{IC13}, it is usual practice to induce auxiliary supervision with new losses \cite{FAN,RobustScanner,DSAN,OOV} appending to the sequential recognition network.
The key of promoting results comes from not only the well-designed module in assistance with the main task  but also the detailed information from additional expensive annotation. For instance, \cite{FAN} handled with the attention drift through an additional focusing attention network aligning the attention center with character-level position annotations. \cite{RobustScanner} designed dual-branch mutual supervision from order segmentation and character segmentation tasks. \cite{DSAN} enhanced the encoding features through an auxiliary branch training the character-level classifier with extra supervision. \cite{OOV} utilized mutual learning of several typical recognition networks, including CTC-based \cite{CRNN}, Attn-based\cite{aster} and segmentation-based \cite{Perspective} methods.
Though effective through auxiliary branches, all these above require the character-level supervision.
Limited by the word-level labels, multi-branch assistance meets much difficulty. Early work \cite{joint} in speech intuitively trained tasks simultaneously by CTC and Attn-based loss but it does not work well in STR as claimed \cite{FAN,GTC}. Thus, GTC \cite{GTC} proposed to enhance CTC-based branch under the guidance from the Attn-based branch, but the Attn-based branch would not be reinforced through the same solution. Among these works\cite{joint,GTC}, mechanized application of shared backbones with multiple losses is adopted, while the relation between the branches, the explicit and implicit tasks, has not been fully exploited.

\section{Methodology}

\subsection{Overall Architecture}

An overview of the proposed architecture is shown in Fig. \ref{figure:architectore_comp}, an end-to-end trainable Reciprocal Feature Learning (RF-L) network.
It aims at exploiting the information from the given word-level annotations and strengthening the feature between different tasks by explicit and implicit annotations.
The network can be divided into four parts: the backbone network, two task branches including text recognition and character counting, and reciprocal feature adaptor interwinding the two branches.
Given an input image, the shared backbone is used to extract visual features from the input images.
Then the character counting branch and text recognition branch utilize the encoded feature for respective tasks simultaneously.
The relation between the two branches is built upon a well-designed module called Reciprocal Feature Adaptor (RF-Adaptor).
It transfers useful messages from one branch to another,
replenishing features in a single task with the high-related partner.
Note that the RF-Adaptor allows bi-directional data flowing and synchronously reinforces the targeted tasks on both sides.

\subsection{Backbone}

The backbone is shared by the two targeted tasks, where any existing feature extractor can be applied in this module.
In these work, we follow the widely-used encoders \cite{wrong} including VGG \cite{VGG}-based \cite{CRNN}, ResNet \cite{ResNet}-based backbone \cite{FAN}.
The feature module (noted as \textit{$F(\cdot)$}) encodes the input image \textit{$x$} with scale \textit{H}$\times$\textit{W} into feature map \textit{V}, where \textit{C}, \textit{$H_{out}$}, \textit{$W_{out}$} denote the output channels, output height and width size.

\begin{equation}
V = F(x), V\in R^{C \times H_{out} \times W_{out}}
\end{equation}
Note that we share shallow layers of backbones, \textit{i.e.,} {the stage-1 and stage-2 in the ResNet-based encoder and the convolution layer ranging from 1 to 5 out of 7 in the VGG-based encoder.} The remained higher layer of features are learned in each branch respectively. We use ResNet\cite{FAN} if not specified in experiments.

\subsection{Character Counting Branch} \label{sec:cnt}

The character counting (short in CNT) branch is responsible for predicting the total character occurrence numbers.
Given the encoded features from the shared backbone, the CNT branch outputs the text length. It's worth mentioning that the supervision for the task can be directly computed from the word-level recognition labels and thus does not require any extra annotation cost.
Similar to \cite{ACE} who predicts the character category-aware occurrence for each text, our prediction of total character counting (a scalar) can be realized through two methods, regression-based prediction with Mean Square Error (MSE) loss, and classification-based with Cross-Entropy loss, respectively as show in Equa. \ref{equ:cnt}.

\begin{equation}\label{equ:cnt}
L_{cnt} = \left\{
\begin{array}{ll}
MSE(\hat{y}_{cnt}, y_{cnt}),              & if \ Regression\\
\\
CrossEntropy(\hat{y}_{cnt},y_{cnt}),             & if \ Classification
\end{array}
\right.
\end{equation}

\noindent where $\hat{y}_{cnt}$ and $y_{cnt}$ are the predicted scalars and the groundtruth (\textit{i.e., the text-length}, like 3 for `cnt') respectively.
{Besides, regarding the unbalanced of character occurring, we count the ratio \textit{$\alpha$} of character appearing numbers, and use \textit{$(1-\alpha)$} as factors to re-weight the loss in training (denoted as \textit{w. Class Balance}), and it works well in both functions in Sec. \ref{exp:cnt}.}

\subsection{Text Recognition Branch} \label{sec:rcg}

The text recognition (short in RCG) branch is typically sequential prediction which can directly use any existing recognition networks \cite{CRNN,aster,DAN,wrong}.

\begin{figure}[t]
	\centering
	\scalebox{0.8}{
		\includegraphics[width=0.95\textwidth]{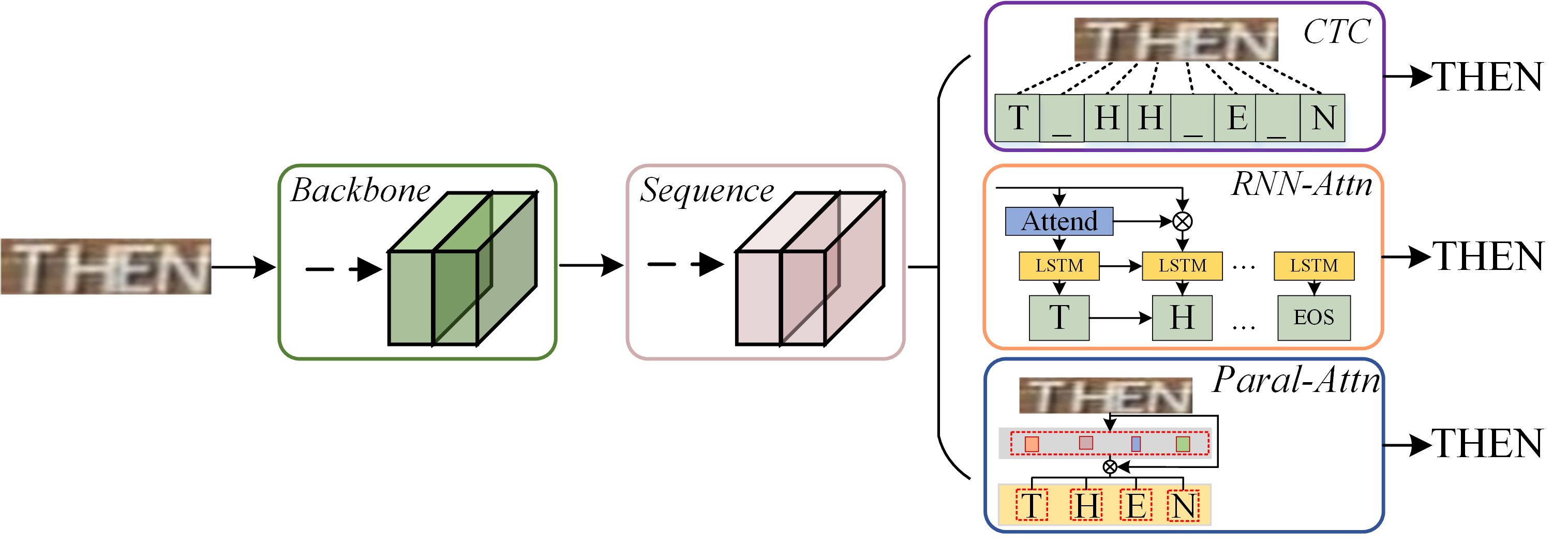}
	}
	\caption{The segmentation-free text recognition networks. }
	\label{fig:seq}
\end{figure}

As shown in Fig. \ref{fig:seq}, the existing segmentation-free recognition models can be mainly grouped into CTC-based \cite{CRNN,2dctc} or Attn-based methods \cite{aster,DAN,wrong}.
The latter group usually has relatively superior performance while the former is more efficient and friendly to real applications.
Specifically, the group of Attn-based decoders include the dominating RNN-based Attn \cite{aster,wrong}, the emerging Parallel Attn \cite{DAN,SRN} (short in Paral-Attn), and Transformer \cite{NRTR}.
As framework of Transformer-based encoder-decoder \cite{NRTR} is largely different from the other CTC or Attn-based pipelines, we do not discuss in this paper.
Therefore,  in this work, we reproduce the CTC \cite{CRNN}, Bilstm-Attn \cite{wrong} (\textit{w.o.} TPS transformation), and Paral-Attn \cite{DAN} models as three representatives among most existing networks.
Note that we are focusing on the feature learning in the recognition networks,
so we adopt mainstreaming models only with the main recognizor \textit{without} any pluggable functional modules (\textit{e.g.,} TPS-based transformation \cite{aster}, high-resolution unit\cite{PlugNet}, semantic enhancement \cite{SRN,SEED1}, position clues \cite{RobustScanner}).
The loss function $L_{rcg}$ in RCG branch is the standard CTC loss or the Cross-Entropy(CE) loss for Attn-based methods.

\subsection{Reciprocal Feature Adaptor}\label{method:adaptor}

\begin{figure}

\centering
\includegraphics[width=0.9\textwidth]{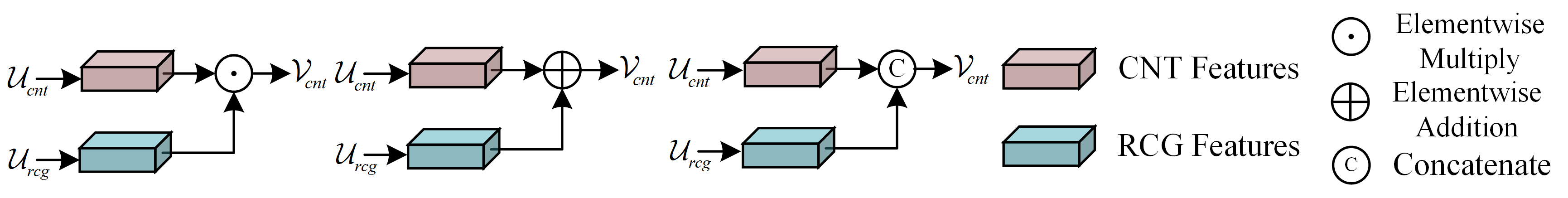}

	\caption{Fusion of the two-branch features in the proposed RF-Adaptor. The figure shows the RCG-to-CNT one-way fusion, and in turn the CNT-to-RCG is similar.}

\label{fuison}
\end{figure}

Instead of the existing works limited by the similartaxis of dual-branch (\textit{e.g.}, CTC and Attn) features \cite{FAN,GTC}, which results from the same supervising labels in word-level STR, we dig the implicit task and labels (\textit{i.e.}, character counting).
We arrange it as an additional branch to train with the recognizor simultaneously, as shown in Fig. \ref{figure:architectore_comp}.
And we are motivated to uncover and thus build the relationship between the two branches.
We believe that a good dual-task mutual assistance majors in two abilities:
{
(1) Offering complementary message from one branch to the other.
(2) Adapting feature from one task to the other.}
Therefore, we propose a module called Reciprocal Feature Adaptor (RF-Adaptor) in order to realize the mutual feature assistance. Specifically, given the separate features from two branches $\mathcal{U}_{cnt}$ and $\mathcal{U}_{rcg}$ as input, the mutual-enhanced features $\mathcal{V} _{cnt}$ and $\mathcal{V}_{rcg}$ via bidirectional adaptors can be formulated as:

\begin{equation}\label{equ:fuse}
\begin{aligned}
\mathcal{V}_{\ast}=\mathcal{U}_{cnt} \diamond \mathcal{U}_{rcg}
\end{aligned}
\end{equation}
where $\diamond$ is the binary operator with two-side inputs, which includes the usual approaches like \textit{element-wise multiplication} (noted in $\odot$),\textit{element-wise addition} (noted in $\oplus$) and \textit{concatenation} (noted in {\footnotesize{$\copyright$}}), as shown in Fig. \ref{fuison}.
In detail, CNT seems inherently contained in RCG, and its supervision offers much less information.
Therefore, RCG can supplement CNT with abundant informative details.
On the other hand, not requiring discriminating the characters, the feature learning in CNT focuses more on the text itself, and thus can obtain more purified representation and not be bothered by the confusing patterns.
For convenience, procedure of generating the enhanced RCG features $\mathcal{V}_{rcg}$ is called \textit{CNT-to-RCG} (short in \textit{C2R}).
Conversely, producing $\mathcal{V}_{cnt}$ is \textit{RCG-to-CNT} (short in \textit{R2C}).
{
\begin{itemize}
\item In \textit{R2C}, as the partner $\mathcal{U}_{rcg}$ contains more information than $\mathcal{U}_{cnt}$, it can replenish the $\mathcal{U}_{cnt}$ a lot via manipulating $\oplus$ on the two features.
\item In \textit{C2R}, $\mathcal{U}_{rcg}$ itself has rich information and $\mathcal{U}_{cnt}$ has more purified features, so CNT plays a role of feature selector or weighting factor for $\mathcal{U}_{rcg}$.
The selector is implemented via a learnable gate, which is designed to suppress the text-unrelated noise and thus refine $\mathcal{U}_{rcg}$ via the $\odot$ operation.

\end{itemize}

Note the direct concatenation $\copyright$ regards the two input equally and thus is not recommended for our asymmetry (with unequal information) dual branches, so we do not apply it in our RF-Adaptor.
Rigorous ablations of the three fusion operations are shown later in Table \ref{ablation:C2R-fuse} with explanations in Sec.\ref{ablation}.

Upon the chosen \textit{fusion} operation $\oplus$ in \textit{R2C} and $\odot$ in \textit{C2R}, and not satisfied by the direct combination on the two-branch features,
we resort to some lightweight modules, which we name them Feature Enhancement (short in \textit{FE}) modules (formulated in $\mathcal{F}(\cdot)$), to dispose of the input $\mathcal{U}_{rcg}$ and $\mathcal{U}_{cnt}$ for better providing the supplementary information to the other. Furthermore, using $\mathcal{I}(\cdot)$ to represent identity mapping, we specify Equa. \ref{equ:fuse} into
\begin{equation}\label{equ:adaptor}
\begin{aligned}
\mathcal{V}_{cnt}&=\mathcal{I}(\mathcal{U}_{cnt}) \oplus \mathcal{F}_{c}(\mathcal{U}_{rcg}) \\
\mathcal{V}_{rcg}&=\mathcal{F}_{r}(\mathcal{U}_{cnt}) \odot \mathcal{I}(\mathcal{U}_{rcg}),
\end{aligned}
\end{equation}

where $\mathcal{F}_{c}(\cdot)$ and $\mathcal{F}_{r}(\cdot)$ are both \textit{FE} modules in \textit{C2R} and \textit{R2C} respectively.
Inspired by the convolutional attention-based modules \cite{SENet,gcnet} which can be easily integrated into the CNNs for feature refinement,
we would like to apply the idea into our setups for a better adaptation on features from one branch to the other.
Differently, instead of self-enhancement as the existing methods \cite{SENet,gcnet} with single input source, we have the dual-branch features from different tasks.
Intuitively, the \textit{FE} modules are designed in two separate ways as details in the following.
\begin{figure}[t]
\centering

\includegraphics[width=0.81\textwidth]{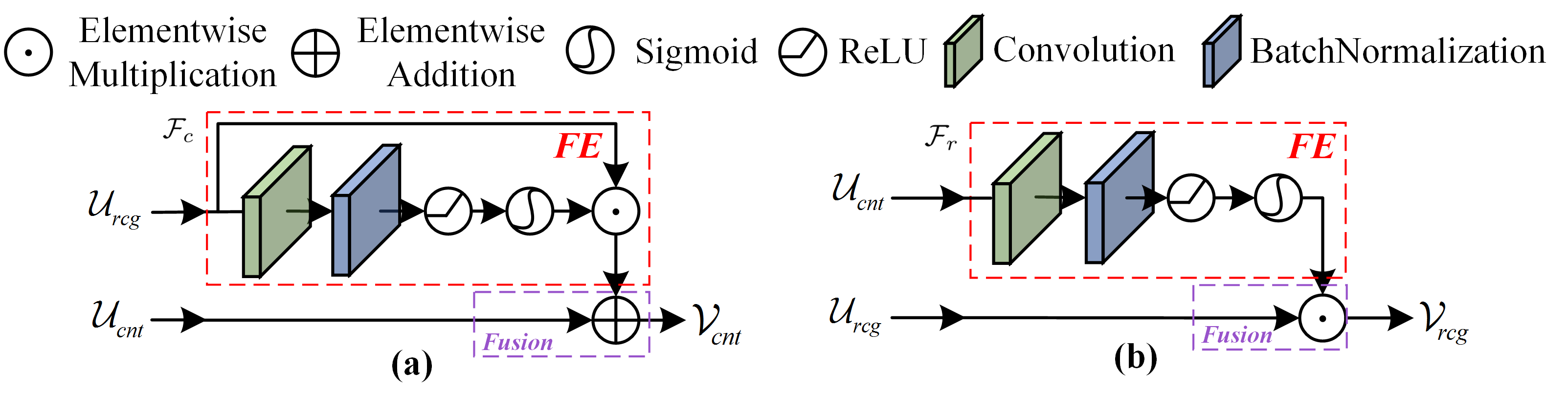}

\caption{The structures for the \textit{FE} blocks. (a) is in \textit{R2C} and (b) is in \textit{C2R}. }

\label{ablation:adaptor}
\end{figure}
\begin{itemize}

\item $\mathcal{F}_{c}(\cdot)$ is the self-reinforced features through multiplying the importance learnt on the context using one 1$\times$1 convolution, Normalization, one ReLU and a sigmoid function sequentially, as Fig. \ref{ablation:adaptor}(a). The function outputs an enhanced RCG features $\mathcal{F}_{c}(\mathcal{U}_{rcg})$ which is expected to suit the CNT branch via end-to-end adaptive learning.
    In the feedforward of \textit{R2C}, producing $\mathcal{V}_{cnt}$ can be represented as the original $\mathcal{U}_{cnt}$, supplemented by useful information from $\mathcal{U}_{rcg}$ via a refinement module on $\mathcal{U}_{rcg}$ itself.
\item $\mathcal{F}_{r}(\cdot)$ is rescaling the feature on each channel per coordinate as Fig. \ref{ablation:adaptor}(b), \textit{i.e.}, using features from $\mathcal{U}_{cnt}$ as soft-attention on the $\mathcal{U}_{rcg}$ to produce an enhanced RCG feature output $\mathcal{V}_{rcg}$.
    In this case, the $\mathcal{F}_{r}(\cdot)$ is similarly composed of one 1$\times$1 convolution, Normalization, ReLU and a sigmoid function orderly. It functions on the $\mathcal{U}_{cnt}$ to compute the importance for each channel and position in the original input RCG features $\mathcal{U}_{rcg}$, which can be seen as using the additional $\mathcal{U}_{cnt}$ for re-weighting the $\mathcal{U}_{rcg}$.
More details and understanding of the proposed RF-Adaptor 
are studied and shown in Section \ref{ablation}. 
\end{itemize}
}

\subsection{Optimization}

The supervision is working on each branch respectively.
The loss function of the overall Reciprocal Feature Learning (RF-L) framework can be formulated by sum of the losses from two branches.

\begin{equation}
	L = L_{cnt} + \lambda L_{rcg}
\end{equation}
where $\lambda$ is 1.0 by default.
$L_{rcg}$ and
$L_{cnt}$ are introduced in Section \ref{sec:cnt} and \ref{sec:rcg}.
The training can be in an end-to-end approach without any pre-training or stage-wise strategies.
Note that we do not adjust the parameters $\lambda$ for a delicate balance, while we merely expect to affirm the mutual assistance on both branches.

\section{Experiments}

\subsection{Datasets}

Following existing benchmarks \cite{wrong}, we train the models on 2 synthetic datasets \textit{MJSynth}(MJ){~\cite{MJ}} and \textit{SynthText}(ST){~\cite{ST}} using word-level annotations.
Note that models are only trained with the above 2 datasets and directly evaluated in the following 7 public testsets without any finetuning stage or tricks.

\textbf{IIIT5K(IIIT)}\cite{IIIT5K} contains 3,000 images of scene texts and born-digital images for evaluation, which are collected from the websites.
\textbf{SVT} \cite{SVT} has 647 images cropped from Google Street View for evaluation.
\textbf{ICDAR2003/2013/2015 (IC03/IC13/IC15)} \cite{IC03,IC13,IC15} offer respectively 860, 867, 1811 images for evaluation, which were created for the Robust Reading Competition in the International Conference on Document Analysis and Recognition(ICDAR) 2003/2013/ 2015.
\textbf{SVTP}\cite{SVTP} contains 645 images for evaluation and the images are collected from Google Street View. Many of the images in dataset are heavily distorted.
\textbf{CUTE80(CT)} \cite{CUTE80} contains 288 cropped images for evaluation. Most of them are curved text images.

\subsection{Tasks and Metrics}
\subsubsection{Scene Text Recognition.}
Text recognition is a popular task in the deep learning community.
Following the existing \cite{wrong}, we regard it as a sequential recognition problem and evaluate the overall performance on the word accuracy.

\subsubsection{Character Counting.}\label{exp:cnt}
Few work uses scene text images for other tasks. 
Noting that Xie et al \cite{ACE} solved the STR problem from a new perspective
which is analogical to our counting task, but they finally transfer the learned feature into sequential outputs as common STR task.
One more step, we explicitly build the task called character counting using the existing STR public datasets.
Annotations for the counting can be directly transferred from the word-level labels, so it does not require extra labor. Given the category-unaware predictions and the counting labels $c_{i}$ for image $\textit{i}$, the RMSE and relRMSE \cite{ACE} are calculated by
\begin{equation}
\begin{aligned}
RMSE &=\sqrt{\frac{1}{N}\sum\nolimits_{i=1}^{N}(\widehat{c_{i}} -c_{i})^2}\\
relRMSE &=\sqrt{\frac{1}{N}\sum\nolimits_{i=1}^{N}{\frac{(\widehat{c_{i}} - c_{i})^2}{c_{i} + 1}}}.
\end{aligned}
\end{equation}
For convenience, we also evaluate the character counting on the ratio of correctly predicted images in ablations.

\subsection{Implementation Details}
All images are resized to 32$\times$100 before inputting to the feature extractor and encoded to the feature map of 1$\times$26. The parameters are initialized using He's method\cite{He} if not specified. AdaDelta optimizer is adopted with the initial learning rate 1. Batch size is set to 128.
The details of backbone are introduced in Sec.3.1. For text recognition, the hidden dimension in RNN and Attn is 512. And the \textit{FE} channel dimension in RF-Adaptor is 512. For quick ablation, we conduct extensive ablations partially on 10\% {MJSynth (MJ)}{~\cite{MJ}} using a simplified Paral-Attn model, where the modules in DAN\cite{DAN} are replaced with linear mapping for Attn decoder.
And the remaining results are following the standard setups\cite{wrong}. All the comparisons are only with word-level annotations. 
\subsection{Ablation Studies} \label{ablation}
\subsubsection{Design of CNT Network.}
We formulate the counting problem as a total number of all-class characters with loss functions designed in two ways, the regression and the Cross-Entropy(CE) loss.
We find it performs better in \textit{regression-based} methods \textit{w. Class Balance} as shown in Table \ref{tab:cnt}, solely in counting tasks\footnote[2]{\textit{Regular} and \textit{Irregular} represent mean accuracy under 4 regular datasets (\textit{i.e.},IIIT5K,SVT and IC03/13) and 3 irregular datasets (\textit{i.e.},IC05,SVTP and CT).}. 
\begin{table}[t]
	\center
	\caption{Comparison of counting results based on Cross-Entropy and regression-based methods, evaluated on the ratio of correctly predicted images (\%).}\label{tab:cnt}
    \scalebox{0.88}{
	\begin{tabular}{|c|cc|cc|}
		\hline
		\multicolumn{1}{|c|}{\multirow{2}{*}{Methods}}&\multicolumn{2}{c|}{\textit{w.o.} Class Balance}&\multicolumn{2}{c|}{\textit{w.} Class Balance}\\
		\cline{2-5}
		            &  Regular\footnote[2] & Irregular &  Regular & Irregular \\
		\hline
		CE          & 89.5 & 78.5 & 93.2 & 83.5 \\
		Regression  & 93.3 & 82.3 & \textbf{94.6} & \textbf{84.5} \\
		\hline
	\end{tabular}
}
\end{table}
\begin{table}[t]
	\caption{Performance of feature fusing in \textit{C2R} and \textit{R2C}. The CNT-to-RCG is evaluated on word accuracy (\%) the same with the RCG task, while the RCG-to-CNT is evaluated on the ratio (\%) of correctly predicted images on the CNT task (similarly in Tab. 3 and Tab. 4). Finally, \textit{C2R} uses $\odot$ (\textit{multiplication}) and \textit{R2C} uses $\oplus$ (\textit{addition}).}\label{ablation:C2R-fuse}
	\center
    \scalebox{0.88}{
	\begin{tabular}{|c|cc|cc|}
		\hline
        \multicolumn{1}{|c|}{\multirow{2}{*}{Methods}}&\multicolumn{2}{c|}{CNT-to-RCG}&\multicolumn{2}{c|}{RCG-to-CNT}\\
        \cline{2-5}
		&  Regular & Irregular &  Regular & Irregular \\
		\hline
		Baseline   & 77.4         & 53.2         & 84.3         & 69.7          \\
		\hline
		$\odot$ Multiplication        & \textbf{79.4(+2.0\%)} & \textbf{54.3(+1.1\%)} & 88.2(+3.9) & 74.6(+4.9)  \\
		$\oplus$ Addition               & 79.0(+1.6) & \textbf{54.3(+1.1\%)} & \textbf{88.5(+4.2\%)} & \textbf{76.6(+6.9\%)} \\
		$\copyright$ Concatenation      & 78.6(+1.2) & 53.1(-0.1) & 88.6(+4.3) & 75.1(+5.4)  \\
		\hline
	\end{tabular}
}
\end{table}
\subsubsection{Design of RF-Adaptor.} \label{sec:abl-rfl}
The design for the cross-branch message learner is focused on \textit{combining two-branch messages}, and \textit{suiting to the other task} as illustrated in Section \ref{method:adaptor}. 
We divide it into 2 steps orderly: (1) Feature Enhancement(\textit{FE}): the attention-based enhancement,
(2) Fusion: the fusion operation.
We do ablation on each point independently for clear explanation on each module in the proposed RF-Adaptor.
\begin{itemize}
\item We first discuss the \textit{fusion} procedure ($\diamond$ in Equa. \ref{equ:fuse}), as it is \textit{necessary} though is the last step in all the three. We compare among the approaches including concatenation, bit-wise multiplication or sum, shown in Fig. \ref{fuison} and Table \ref{ablation:C2R-fuse}.
\item The proposed \textit{FE} modules are inspired from the channel attention\cite{gcnet} and broaden it into the two-branch input networks. Comparing RF-Adaptor and RF-Adaptor \textit{w.o. FE} in Table \ref{tab:adaptor}, it shows the effectiveness of the \textit{FE} module not only in \textit{C2R} but also in \textit{R2C} direction.
\end{itemize}
\subsection{Comparison with State-of-the-art Approaches}\label{sec:sota}
\subsubsection{Scene Text Recognition.}
\begin{figure}[t]
\centering
\includegraphics[width=0.85\textwidth]{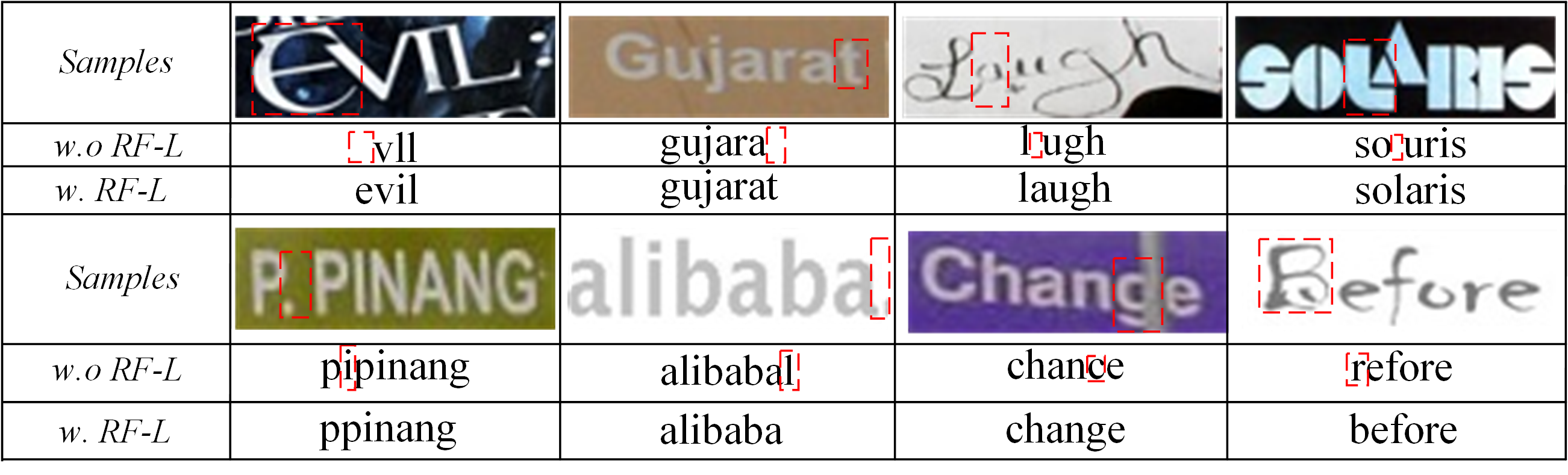}
\caption{Visualizations on STR tasks in comparison with the model \textit{w.} and \textit{w.o RF-L}.}
\label{fig:vis}
\end{figure}

In order to verify the generalization of the proposed approach, we evaluate our proposed RF-L in form of two networks based on Attention,
the strong baseline Bilstm-Attn~\cite{wrong} and DAN~\cite{DAN} based on the well-designed decoupled attention.
\begin{table}[t]
	\caption{Effect of different components in RF-Adaptor.}\label{tab:adaptor}
	\center
	\scalebox{0.8}{
		\begin{tabular}{|c|cc|cc|cc|}
			\hline \multicolumn{1}{|c|}{\multirow{2}{*}{Methods}}&\multicolumn{2}{c|}{Modules}&\multicolumn{2}{c|}{CNT-to-RCG}&\multicolumn{2}{c|}{RCG-to-CNT}\\
			\cline{2-7}
			&Fusion&\textit{FE} &  Regular & Irregular &  Regular & Irregular \\
			\hline
			Baseline &&                    & 77.4         & 53.2         & 84.3         & 69.7  \\
			RF-Adaptor \textit{w.o.}\textit{FE}    &$\surd$&        & 79.4(+2.0) & 54.3(+1.1) & 88.5(+4.2) & 76.6(+6.9) \\

			RF-Adaptor          &$\surd$&$\surd$&\textbf{79.6(+2.2)} & \textbf{55.4(+2.2)} & \textbf{90.4(+6.1)} & \textbf{77.0(+7.3)} \\
			\hline
		\end{tabular}
	}
\end{table}
\begin{table}[t]
	\centering	
    \caption{Comparison with SOTA recognition models with their training settings. We only report the results using MJ and ST without extra data or annotations. The best accuracy is in \textbf{bold} and the second one is denoted in \underline{underline}.}\label{tab:sota}
    \scalebox{0.81}{
	\begin{tabular}{|l|c|c|ccccccc|cc|}
		\hline
		\multicolumn{1}{|c|}{\multirow{2}{*}{Methods}}& \multicolumn{1}{c|}{\multirow{2}{*}{Year}} & \multicolumn{1}{c|}{\multirow{2}{*}{Training data}} & \multicolumn{7}{c|}{\multirow{1}{*}{Benchmark}} & \multicolumn{2}{c|}{\multirow{1}{*}{Avg. Acc}}                                                                                                                                           \\
\cline{4-12}
		&                          &                                & IIIT                  & SVT                      & IC03                     & IC13                     & IC15                  & SVTP                  & CT  & Regular & Irregular                 \\ \hline

		CRNN \cite{CRNN}                     & 2016                      & MJ                             & 78.2                  & 80.8                     & 89.4                     & \multicolumn{1}{c}{-}                    & \multicolumn{1}{c}{-} & \multicolumn{1}{c}{-} & \multicolumn{1}{c|}{-} & \multicolumn{1}{c}{-} & \multicolumn{1}{c|}{-}\\

		AON \cite{AON}   & 2018                   & MJ+ST                     & 87.0                  & 82.8                     & 91.5                     &  \multicolumn{1}{c}{-}                     & \multicolumn{1}{c}{-}   & 73.0                  & 76.8 & \multicolumn{1}{c}{-} & \multicolumn{1}{c|}{-}   \\
		
		NRTR \cite{NRTR}   & 2018               & MJ+ST                   & 90.1                 & \textbf{91.5}                    & {94.7}& \multicolumn{1}{c}{-}                    &  79.4                          & \textbf{86.6}                 &  80.9                 &   \multicolumn{1}{c}{-}   & \underline{82.3} \\
		
		ASTER \cite{aster}                    & 2019                   & MJ+ST                          & 93.4                 & 89.5                     &  94.5 & 91.8                    & -                  & 78.5                   & 79.5 & 92.3 & \multicolumn{1}{c|}{-}             \\
		TPS-Bilstm-Attn \cite{wrong} & 2019           & MJ+ST                      & 87.9  &
		87.5                       & \underline{94.9}              &      93.6              & 77.6    & 79.2 &  74.0  & 91.0 & 76.9 \\
		
		
		AutoSTR \cite{AutoSTR}$\ast$                     & 2020                   & MJ+ST                          &  \underline{94.7}                 & \underline{90.9}                   &  93.3                        & \underline{94.2}                      & \underline{81.8}                  & 81.7                & \multicolumn{1}{c|}{-}     &  \textbf{93.2} & \multicolumn{1}{c|}{-}         \\
		
		RobustScanner \cite{RobustScanner}$\dagger$                    & 2020                   & MJ+ST                          & \textbf{95.3}                  & 88.1                &  \multicolumn{1}{c}{-}                       & \multicolumn{1}{c}{-}                      &   \multicolumn{1}{c}{-}                & 79.5                & \textbf{90.3}    & \multicolumn{1}{c}{-}  & \multicolumn{1}{c|}{-}                \\  \hline
		
		Bilstm-Attn \cite{wrong}\footnote[3]     & 2019                      & MJ+ST                          &  93.7 & 89.0 & 92.3 & 93.2 & 79.3 & 81.2 & 80.6     & 92.1 & 80.4                \\
		
		Bilstm-Attn \textit{w.} RF-L                & -                      & MJ+ST                          & 94.1                      & 88.6                        &  \underline{94.9}                      &   \textbf{94.5}                      & \textbf{82.4}                      &  {82.0}                   &  82.6      & \underline{93.0}(+0.9) & \textbf{82.4}(+2.0)                \\
		DAN \cite{DAN}\footnote[4] & 2020   & MJ+ST      & 93.4 & 87.5 & 94.2 & 93.2 & 75.6 & 80.9 & 78.0  &92.1&78.2\\
		DAN  \textit{w.} RF-L & -& MJ+ST     & 94.0 & 87.7 & 93.6 & 93.5 & 76.7 & \underline{84.7} & 77.8  &92.2(+0.1) &79.7(+1.5)\\
		
    \hline
	\end{tabular}
}
	\label{table:3}
\end{table}
For fair comparison, we train the baseline using the released code\footnote[3]{We use the code at https://github.com/clovaai/deep-text-recognition-benchmark.} and resource\footnote[4]{We use the code and pretrained parameters released at https://github.com/Wang-Tianwei/Decoupled-attention-network.} from authors under the same experiment setups, without additional dataset, augmentation, rectification, stage-wise finetuning or other training tricks.
As Bilstm-Attn is the solely recognition networks, we supplement the task with CNT branch as auxiliary via RF-L. The comparisons between the SOTA approaches are reported in Table \ref{tab:sota}.
It shows that the strong baseline can be further promoted through multi-task without extra annotation.
Bilstm-Attn w. RF-L obtains the obvious improvement on the overall 7 benchmarks, especially over 2\% above the original on IC03, IC15 and CT.
Visualizations are as Fig. \ref{fig:vis}. It can be seen that the missing or the redundant characters
can be effectively corrected via the RF-L, \textit{e.g.}, the word `evil', `alibaba'. Besides, the images with distraction like `change' can be easier to recognize through the enhanced features.
It is similar when equipped to DAN where the promotion is also obvious compared to the original one.
{
Note that our results are only slightly overwhelmed on 2 out of 7 datasets by recent AutoSTR\cite{AutoSTR}(with $\ast$) and RobustScanner \cite{RobustScanner}(with $\dagger$).
The former applies backbones specific for each dataset via automated network search, which relies on validation data in all 7 benchmarks.
The latter uses specific pre-processing on the input and evaluate on the best of 3-sibling image group.
}
\begin{table}[tb]
\caption{Performance of character counting with existing methods in evaluation of RMSE/relRMSE (smaller is better). The best results for each test set are in \textbf{bold}.}\label{tab:sota-cnt}
\center
\scalebox{0.88}{
\begin{tabular}{|c|cc|cc|cc|cc|}
\hline
\multicolumn{1}{|c|}{\multirow{2}{*}{Methods}}   &  \multicolumn{2}{c|}{ACE \cite{ACE}\footnote[5]} & \multicolumn{2}{c|}{ACE w. RF-L} & \multicolumn{2}{c|}{CNT} & \multicolumn{2}{c|}{CNT w. RF-L} \\
\cline{2-9}
              &  RMSE & relRMSE  & RMSE & relRMSE & RMSE & relRMSE & RMSE & relRMSE \\
\hline
IIIT          & 0.477 &  0.169   & 0.323 &  0.133  & 0.300  &   0.128  & \textbf{0.272} &  \textbf{0.115}   \\
SVT           & 0.963 &  0.361   & 0.890 &  0.326  & 0.455  &   0.165  & \textbf{0.455} &  \textbf{0.164}   \\
IC03          & 0.555 &  0.206   & 0.509 &  0.192  & 0.372  &   0.147  & \textbf{0.352} &  \textbf{0.138}   \\
IC13          & 0.518 &  0.193   & 0.502 &  0.188  & 0.275  &   0.107  & \textbf{0.268} &  \textbf{0.106}   \\
IC15          & 0.889 &  0.364   & 0.896 &  0.361  & 0.614  &   0.261  & \textbf{0.604} &  \textbf{0.256}   \\
SVTP          & 1.389 &  0.499   & 1.414 &  0.514  & \textbf{0.724}  &   0.258  & 0.747 &  \textbf{0.256}   \\
CT            & 1.001 &  0.443   & 1.200 &  0.442  & 0.854  &   0.420  & \textbf{0.835} &  \textbf{0.368}   \\
\hline
\end{tabular}
}
\end{table}
\subsubsection{Character Counting.}
In scene text images, character counting task has not been introduced. We first build the benchmark using the same public 7 test datasets as the traditional STR tasks. Similarly, we also use the 2 synthetic datasets for training and evaluate directly (without finetuning) on these non-homologous test sets. Among existing works in STR research field, no former results have been reported according to this task. Only the approach \cite{ACE} has something related, partly belonging to the scope the counting tasks but differing in the evaluation. Therefore, we re-implement the ACE\cite{ACE} with the release\footnote[5]{As no reported character counting results, we re-implement the model using the released code at https://github.com/ summerlvsong/Aggregation-Cross-Entropy.} and train models using the same backbone and training setups with our proposed approach.
Table \ref{tab:sota-cnt} shows the comparison, which verifies the effectiveness of our proposed CNT branch and the promotion for CNT task from the other (\textit{i.e., RCG}) branch through the proposed RF-Learning.
\section{Discussions}
\subsection{Effectiveness of RF-Learning Framework}
\subsubsection{Mutual Promotion of Explicit and Implicit Tasks.}\label{sec:dis-mutual}
In Table \ref{dis:1}, given the explicit text recognition task and the implicit character counting task, we intuitively use the shared features for joint learning (denoted in \textit{JT-L}) as a stronger baseline than the single RCG or CNT.
Comparing \textit{RCG} and all kinds of \textit{RCG w.CNT}, the assistance from CNT is obvious, where the gain on average accuracy ranges from +1.1\% (\textit{JT-L}) to +2.0\% (\textit{Bidirectional RF-L}). It is similar in turn for the RCG's effects on the CNT task.
Even the simple \textit{JT-L} with shared backbone can help a lot.
We attribute it to the additional supervision though implicit labels contained in the main task from a different optimization goal.
Conversely, the assistance from RCG to CNT tasks is similar.
\begin{table}[t]
	\caption{Performance of RCG tasks and CNT tasks, evaluated on word recognition accuracy (\%) and characters counting accuracy of total images (\%) respectively.}\label{dis:1}
    \center
	\scalebox{0.78}{
    \begin{tabular}{|c|cc|cc|ccccccc|c|}
        \hline
        \multicolumn{1}{|c|}{\multirow{2}{*}{Methods}} & \multicolumn{2}{c|}{Branch}& \multicolumn{2}{c|}{Direction} & \multicolumn{7}{c|}{Benchmark}& Avg.\\
        \cline{2-12}		
		 &  RCG & CNT & C2R & R2C & IIIT & SVT & IC03 & IC13 & IC15 & SVTP & CT & Acc \\
		\hline
		RCG &$\surd$ &&&&90.0 & 82.8 & 87.6 & 89.0 & 72.4 & 71.0 & 73.3 & 81.3 \\
		RCG w. CNT (JT-L) &$\surd$ &$\surd$&&&89.6 & 83.9 & 92.6 & 91.7 & 72.6 & 74.0 & 78.1 & 82.4(+1.1) \\
        RCG w. Fixed CNT (RF-L) &$\surd$ &$\surd$&$\surd$&& 90.2 & 86.7 & 92.2 & 91.6 & 73.2 & 76.0 & 79.5 & 82.8(+1.5)  \\
		RCG w. CNT (Unidirectional RF-L) &$\surd$ &$\surd$&$\surd$&& 90.7 & 86.6 & 92.6 & 91.2 & 73.2 & 76.0 & 80.2 & 82.9(+1.7) \\
		RCG w. CNT (Bidirectional RF-L) &$\surd$ &$\surd$&$\surd$&$\surd$&90.3 & 85.8 & 92.2 & 93.0 & 73.8 & 75.8 & 77.8 & 83.3(+2.0) \\
		\hline
		CNT  & &$\surd$&&&92.5 & 93.0 & 96.3 & 95.6 & 84.2 & 85.0 & 85.8 & 89.4 \\
		CNT w. RCG (JT-L) &$\surd$ &$\surd$&&&93.0 & 94.3 & 96.2 & 96.1 & 84.9 & 86.4 & 83.7 & 89.8(+0.4) \\
        CNT w. Fixed RCG (RF-L) &$\surd$ &$\surd$&&$\surd$& 91.6 & 92.9 & 96.5 & 96.0 & 86.0 & 87.3 & 87.2 & 89.9(+0.5)  \\
		CNT w. RCG (Unidirectional RF-L) &$\surd$ &$\surd$&&$\surd$&92.6 & 93.5 & 96.6 & 95.2 & 86.0 & 86.7 & 89.6 & 90.0(+0.6) \\
		CNT w. RCG (Bidirectional RF-L) &$\surd$ &$\surd$&$\surd$&$\surd$& 93.5 & 94.0 & 96.7 & 95.7 & 85.5 & 86.7 & 88.9 & 90.3(+0.9) \\
        \hline
	\end{tabular}
    }
\end{table}
\subsubsection{Advantage of RF-Adaptor Module.}
We have chosen the RF-Adaptor structure in Section \ref{sec:abl-rfl}.
Here we compare different training strategies using the RF-Adaptor.
We compare 3 approaches as (1)\textbf{Fixed CNT(RF-L)}: the fusion (in RF-Adaptor) is with pre-fixed encoded features from the trained independent CNT network (in the preceding order), without end-to-end adaptive adjusting procedure, (2)\textbf{Unidirectional RF-L}: only CNT-to-RCG one-pass message flow is applied and
(3)\textbf{Bidirectional RF-L}: the end-to-end bidirectional interactive training with the two branches is carried on.
All of the three types achieve better results than the \textit{CNT(JT-L)}, which verifies the design of the RF-Adaptor module is truly beneficial in combining the two-branch features.
Among these, the \textit{Bidirectional RF-L} performs above the others by a large margin, around 0.9\% average gain on 7 benchmarks above the traditional joint learning and 2.0\% gain compared to the original single RCG.
Similar effect can be also found in the CNT task. The structure design, the bidirectional flowing and the end-to-end dual-task optimization all contribute to the effectiveness of the RF-L.
\subsection{Generalization of the Proposed Approach}
\subsubsection{Backbones and Decoders.}
We excavate exhaustively the mutual relation between the explicit (text recognition) task and the commonly-ignored implicit (character counting) task.
We regard the proposed RF-L a general mechanism, and a plug-and-in tools for different settings.
For the network-unawareness, we supplement CNT branch with different kinds of backbones, including VGG\cite{CRNN}, and ResNet\cite{FAN}, and different decoders including CTC \cite{CTC}, RNN-based Attn \cite{aster}, 
shown in Table \ref{backbone}.
All these variable networks, regardless of changing backbones or decoders conformably show the effectiveness and good adaptation of RF-L.
\begin{table}[t]

	\caption{Performance of recognition tasks with the proposed CNT branch through reciprocal-feature learning with different reproduced recognition mainstreaming networks, including variable backbones and decoders.}\label{backbone}

	\center
    \scalebox{0.9}{
	\begin{tabular}{|cc|c|ccccccc|c|}
		\hline
		Encoder& Decoder & w. CNT (RF-L) &IIIT & SVT & IC03 & IC13  & IC15 & SVTP & CT & Avg.Gain \\
		\hline
		VGG &Bilstm-Attn&& 91.2 & 85.5 & 92.6 & 92.1 & 77.5 & 77.7 & 73.6 & \\
		VGG &Bilstm-Attn&$\surd$ & 91.8 & 86.9 & 92.9 & 92.9 & 78.0 & 78.9 & 74.7 & +0.9 \\
        \hline
		ResNet  &Bilstm-Attn&       & 93.7 & 89.0 & 92.3 & 93.2 & 79.3 & 81.2 & 80.6 & \\
		ResNet&Bilstm-Attn& $\surd$ & 94.1 & 88.6 & 94.9 & 94.5 & 82.4 & 82.0 & 82.6 &+1.4\\
		\hline
		ResNet  &CTC &        & 91.7 & 85.8 &	91.5 &	91.7 &	74.1 &	73.2 &	76.7 &  \\
		ResNet  &CTC & $\surd$ & 92.1 &	86.9 &	92.1 &	92.4 &	76.5 &	75.8 &	78.9  &+1.5\\
		\hline
	\end{tabular}
}
\end{table}

\begin{table}[t]
	\caption{Performance of the proposed framework adapted to ACE in evaluation of both word recognition accuracy and character counting performance evaluated in RMSE.}\label{tab-1}
    \center
    \scalebox{0.9}{
	\begin{tabular}{|c|cc|cccc|cccc|}
		\hline
        \multicolumn{1}{|c|}{\multirow{3}{*}{Methods}}&\multicolumn{2}{c|}{Auxiliary} &\multicolumn{4}{c|}{RCG Accuracy (\%)} & \multicolumn{4}{c|}{CNT RMSE}\\
        \cline{2-11}
        & CNT & RCG &  IIIT  & SVT  & IC03  & IC15 &  IIIT  & SVT  & IC03  & IC15 \\
        \cline{8-11}
        \hline
		ACE      & &           & 87.5  & 81.8 & 89.9 & 67.5  & 0.477 & 0.963 & 0.555& 0.889 \\
		w. RCG (RF-L) & &$\surd$& \textbf{88.4}  & \textbf{83.8} & 90.2 & 70.0 & \textbf{0.323} & 0.890 & 0.518 & 0.896  \\
	    w. CNT (RF-L) &$\surd$& & \textbf{88.4}  & 83.6 & \textbf{90.3} & \textbf{70.1} & {0.327} &  \textbf{0.886}& \textbf{0.514}  & \textbf{0.884} \\
		\hline
	\end{tabular}
}
\end{table}

\subsubsection{ACE-Related Task Formulations.}
We go deeper into the proposed framework with a far more interesting experiment.
The aforementioned are from the perspective of recognition or the new-proposed character counting tasks,
but we observe an intermediate zone where ACE \cite{ACE} solved the sequential recognition task.
Compared to the normal RCG tasks, it optimizes the loss function ignoring the ordering of character occurrence and serialize the output to the final string.
While compared to CNT tasks, it pays more attention on category-aware character appearance. It also means, we can regard the ACE-based network in the perspective of auxiliary assistance to both RCG and CNT tasks. Applied in our proposed RF-L framework, we use it as the RCG branch assisted by our CNT branch, and conversely implement it as the CNT as auxiliary branch to our major RCG task.
Table \ref{tab-1} shows that despite the role the ACE network plays in our framework, it can be enhanced,
so does the other concurrent branch. 

\section{Conclusion}

\label{conclusion}
In this paper, we excavate the implicit tasks in the traditional STR, and design a two-branch reciprocal feature learning framework.
Through exploiting the complementary effect between explicit and implicit tasks, \textit{i.e.}, text recognition and character counting, the feature is reliably enhanced.
Extensive experiments show the effectiveness of the proposed methods both in STR tasks and the counting tasks, and the reciprocal feature learning framework is easy to equip with variable networks and tasks.
We offer abundant ablation studies, generalizing experiments with deeper understanding on the tasks.

\bibliographystyle{splncs04}
\bibliography{bibfile}
\end{document}